\documentclass[sigconf]{acmart}

\usepackage{balance}
  
\usepackage{array}

\def\BibTeX{{\rm B\kern-.05em{\sc i\kern-.025em b}\kern-.08emT\kern-.1667em\lower.7ex\hbox{E}\kern-.125emX}}
    
\copyrightyear{2019} 
\acmYear{2019} 
\setcopyright{acmlicensed}
\acmConference[KDD '19]{The 25th ACM SIGKDD Conference on Knowledge Discovery and Data Mining}{August 4--8, 2019}{Anchorage, AK, USA}
\acmBooktitle{The 25th ACM SIGKDD Conference on Knowledge Discovery and Data Mining (KDD '19), August 4--8, 2019, Anchorage, AK, USA}
\acmPrice{15.00}
\acmDOI{10.1145/3292500.3330693}
\acmISBN{978-1-4503-6201-6/19/08}

\begin{document}

\title{E.T.-RNN: Applying Deep Learning to Credit Loan Applications}

\author{Dmitrii Babaev}
\email{dmitri.babaev@gmail.com}
\affiliation{
  \institution{Sberbank AI Lab}
}

\author{Maxim Savchenko}
\email{savvvan@gmail.com}
\affiliation{
  \institution{Sberbank AI Lab}
}

\author{Alexander Tuzhilin}
\affiliation{
  \institution{New York University}
}
\email{atuzhili@stern.nyu.edu}

\author{Dmitrii Umerenkov}
\email{d.umerenkov@gmail.com}
\affiliation{
  \institution{Sberbank AI Lab}
}

\date{30 July 1999}

\begin{abstract}
In this paper we present a novel approach to credit scoring of retail customers in the banking industry based on deep learning methods. We used RNNs on fine grained transnational data to compute credit scores for the loan applicants. We demonstrate that our approach significantly outperforms the baselines based on the customer data of a large European bank. We also conducted a pilot study on loan applicants of the bank, and the study produced significant financial gains for the organization.
In addition, our method has several other advantages described in the paper that are very significant for the bank.
\end{abstract}



\maketitle

\section{Introduction}

Credit scoring is a very important problem for the banking industry because of the huge financial implications for the banks. Banking industry has developed credit scoring models ever since the middle of the XX century  and has perfected these models ever since, investing millions of dollars in this process. Traditional credit scoring models rely on a loan application questionaire, applicants credit history and various other aggregate financial information relevant to the customer's application. These models use traditional machine learning methods, such as logistic regression, to compute the credit score of a customer that is indicative if the customer would return the loan or not. Although widespread and useful, these models have certain limitations. First, credit scoring require extensive feature engineering and deep domain knowledge in order to design good features.
Second, if the customer does not have significant credit history, it is hard to make reliable scoring decisions regarding that person. Third, the currently existing models do not take full advantage of all the data available about the customer in modern settings.

In this paper we propose a novel approach, called \textit{Embedding-Transactional Recurrent Neural Network (E.T.-RNN)}, to compute credit scores of the bank customers by examining history of their credit and debit card transactions.
We do it using a deep learning (DL) approach, as opposed to more traditional machine learning methods.
Note that this approach is applicable only to those customers that have credit or debit cards with the bank. Since a significant percent of the applicants indeed have credit or debit cards, our method works for a large segment of the applicants. Furthermore our proposed method has the following advantages in comparison to the current credit scoring methods.
First, as is shown in the paper, the proposed DL-based method outperforms the baselines, including models currently in use in the bank, resulting in significant financial gains.
Second, the proposed DL-based model works directly on the customer transactions and does not need extensive feature engineering requiring deep domain expertise (generating hundreds or thousands of hand-crafted aggregate features).
Third, our model works exclusively on the transactional data and, therefore, does not require any additional input from the client. This means that we can make credit loan decisions very fast, ideally in near real-time, because the whole credit scoring process is fully automated.
Fourth, information in the transactional data is exceptionally hard to forge. Hence there is no need to check the correctness of the data, unlike the questionnaire and some other data sources used for scoring.
Fifth, even the customers without any credit history can be assessed for credit worthiness, their transactional history constituting a source for estimating credit risks.
Finally, the proposed method  constitutes a \textit{fair} approach to credit scoring, as it does not use information about the individual and therefore cannot be used to discriminate the credit applicants by various demographic factors.
These constitute very significant advantages vis-a-vis current loan practices and have a potential to disrupt the retail banking loan industry.

One issue with the proposed approach constitutes the interpretability of the black-box models, such as neural networks.
Different organizations across the world have different philosophies regarding applying black-box models to credit scoring problems. While in some countries lack of interpretability is considered as clear "no-go", in other parts of the world it is considered more appropriate to use such models for the credit scoring tasks. Also, there has been significant progress in solving the black box interpretation problems in the last few years \cite{DBLP:journals/corr/ChoiBSSS16}, \cite{gupta2018lisa}, \cite{mccoy2018rnns} and we expect this progress to accelerate even faster moving forward. Therefore, we believe that the issue of using the black-box models for credit scoring will be less relevant in the future.

This paper makes the following contributions.
First, we propose to use neural networks on the customers fine-grained transactional data for credit scoring applications in the banking industry.
Second, we tested our method against the benchmarks on the historical data and achieved superior performance.
Third, we conducted a pilot study on loan applicants and produced significant financial gains for the bank.

The rest of the paper is organised as follows. In Section 2 we discuss the related work. In Section 3 we describe the proposed method. In Section 4 we present our experiments and in Section 5 the results of these experiments. Section 6 and 7 are dedicated to the discussion of our results and conclusions.

\section{Related work} \label{sec-rw}

There is a large amount of research on credit scoring problems for the banking industry going back to first half of the XX century \cite{NBERc12952}. A wide range of methods has been used for this task, including logistic regression \cite{RePEc:cup:jfinqa:v:15:y:1980:i:03:p:757-770_00}, decision trees \cite{makowski1985credit}, boosting \cite{Bastos2008CreditSW}, support vector machines (SVM) \cite{HUANG2007847} and neural networks (NN) \cite{west2000neural}. Credit scoring methods historically relied on using questionnaire data and applicant's credit history. However new data sources have been utilized more recently to increase scoring quality by using telecom data \cite{bjorkegren2017behavior} and transactional data \cite{khandani2010consumer}, \cite{bellotti2013forecasting}, \cite{KVAMME2018207}, \cite{chi2012hybrid}, \cite{RePEc}.

Most of the previous approaches to credit scoring used {\em aggregated} transactional data either globally \cite{chi2012hybrid} or over some time window, such as a month \cite{khandani2010consumer}, \cite{bellotti2013forecasting} or a day \cite{KVAMME2018207}, and most of them relied on the classical ML methods. For example, in \cite{khandani2010consumer} authors used generalized classification and regression trees on monthly transactional statistics. In \cite{bellotti2013forecasting} authors used discrete survival models on monthly transactional statistics. Furthermore, some authors used NN-based approaches to credit scoring on the aggregated transactional data. For example in \cite{KVAMME2018207} authors applied shallow convolutional neural networks on daily transactional statistics.

Furthermore, \cite{RePEc} has developed some credit scoring models on the {\em unaggregated} transactional data. However, they used classical ML methods, such as SVMs and weighted-vote relational neighbour classifiers, in their models. Moreover, they focused on the connectivity problem in their work to estimate credit risk and used only information of who transacted with whom, without deploying the full power of the transactional data.

Also, NNs have been applied to the analysis of the transactional data, but in other types of applications. In \cite{fraud_lstm} authors used Long Short Term Memory (LSTM) Recurrent Neural Network (RNN) \cite{gers1999learning} on individual transaction features for detection of fraudulent transactions. For a review of NNs methods in credit card fraud detection see \cite{abdallah2016fraud}. In \cite{zhang2017credit}  authors applied LSTM RNN for predicting credit scores for a peer-to-peer lending platform.

The main contribution of this work is that we use Neural Network methods for traditional {\em banking credit scoring} problems on the {\em unaggregated} transactional data.
In this paper, we use an RNN based method in the credit scoring problem. We describe our approach to this problem in the next section.

\section{The method}

\subsection{Transactional data}

Our method computes credit scores using transactional data,
each client having multiple credit card transactions, and each transaction having several attributes, both categorical and numerical, and occurring at a certain time. Our data can be described as multivariate time-series data, the schema of which is presented in Table \ref{tab-tr-data}. Merchant type field represents the kind of a merchant, such as airline, hotel, restaurant, etc. (note that it is impossible to restore the real merchant organization identifier from this field).

\begin{table}[ht]
\caption{Data structure for a single client}
\begin{tabular}{ | m{7em} |  m{5em} m{5em} m{5em}| }
\hline
\textbf{Amount} & 230 & 5 & 40 \\
\textbf{Currency} & EUR & USD & USD \\
\textbf{Country} & France & US & US \\
\textbf{Time} & 16:40 & 20:15 & 09:30 \\
\textbf{Date} & Jun 21 & Jun 21 & Jun 22 \\
\textbf{Merchant Type} & Restaurant & Transport\-ation & Household Appliance \\
\textbf{Card type} & Visa Classic & Visa Classic & Visa Gold \\
\textbf{Issuing} & 90/10735 & 90/01735 & 90/01779 \\
\textbf{Branch} &&& \\
\textbf{N opened} & 1 & 1& 1 \\
\textbf{debit cards} &&& \\
\textbf{N opened} & 1 & 1& 1 \\
\textbf{credit cards} &&& \\
\hline
\end{tabular}
\label{tab-tr-data}
\end{table}

In the next section we describe an architecture of the neural network that computes credit scores using that transactional data.

\subsection{Architecture overview}

RNNs are used for processing sequential information.  In a way, RNNs have "memory" over previous computations and use information from the previous time-steps in addition to the current input in order to produce next output. This approach is naturally suited for many NLP tasks including text classification, machine translation and language modelling \cite{mikolov2010recurrent}.

Our \textit{Embedding-Transctional RNN (E.T.-RNN)} architecture is presented in Figure \ref{fig-arch} and is inspired by the NLP methods in the context of deep learning \cite{mikolov2010recurrent}. We treated the credit scoring task as a text classification task, using clients as texts and transactions as individual words.

As Figure \ref{fig-arch} shows, the E.T.-RNN model consists of three parts: embedding layers, recurrent encoder and classifier. We will explain each of them in the rest of this section. Note that all the parts are trained \textit{simultaneously} in the end-to-end manner.

\subsubsection{Embeddings}

Credit card transactions are mapped into a latent space before being passed to the encoder RNN. In particular, each categorical variable in each transaction is encoded to a low-dimensional vector via a corresponding embedding layer. The embedding layers are randomly initialized and trained simultaneously with the encoder. We have treated the timestamp as a collection of categorical variables each representing a date part (hour, weekday, month). Each transaction is represented as a concatenation of scalar variables and embeddings of categorical variables.

\subsubsection{Encoder}

We used a single layer RNN based on Gated Recurrent Unit (GRU) \cite{DBLP:journals/corr/ChoMGBSB14} as an encoder.  The hidden vector from the last time step was used as the representation of the client. Note that this approach is also commonly used for text analysis \cite{Sutskever:2014:SSL:2969033.2969173}.

\subsubsection{Classifier}

The hidden vector from the last time step is finally passed to the fully connected classifier sub-network.
It turned out that a simple linear classifier outperformed several alternative approaches in our experiments and therefore we used it in our architecture.

\begin{figure}[ht]
  \caption{Final architecture}
  \includegraphics[width=0.46\textwidth]{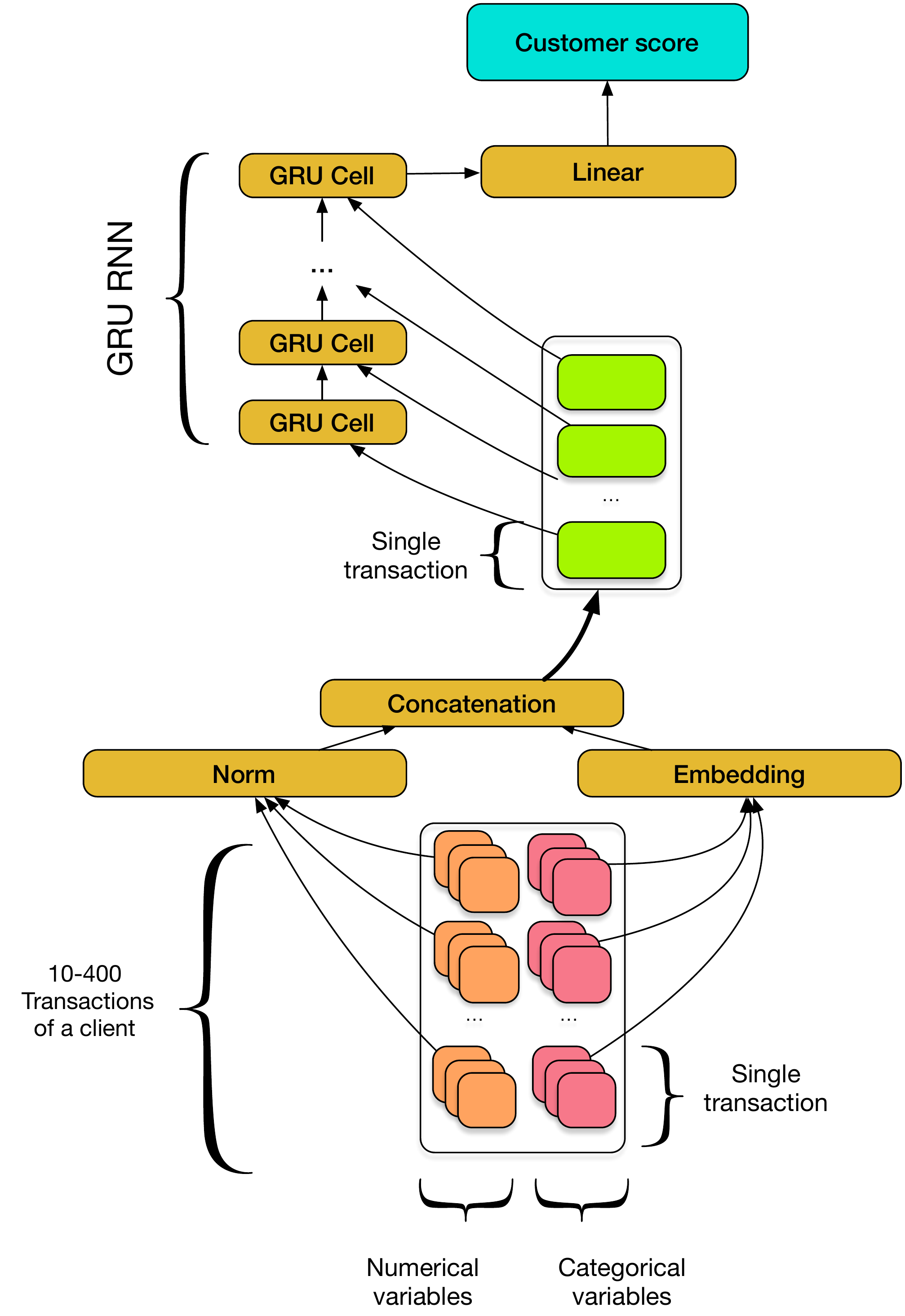}
  \label{fig-arch}
\end{figure}

More generally, we experimented extensively with different types of deep learning architectures, as explained in Section \ref{sec-arc-sel}, and the architecture presented in Figure \ref{fig-arch} turned out to be the best for our experiments.

\subsection{Loss function}

In this work we use the standard area under the ROC curve (ROC AUC) performance measure.

Several loss functions can be used as a proxy for the task of maximising ROC AUC, including the classic binary cross-entry loss: $L_{CE}(p, y) = - \sum_i y_ilog(p_i)$ and margin ranking loss: $ L_R(p_1, p_2, y) = \max(0, -y * (p_1 - p_2) + margin $ which directly optimizes ROC AUC.

In the final version of our model we decided to use margin ranking loss with margin 0.01, which showed the best results on our data, as presented in Section \ref{sec-arc-sel}.

\subsection{Ensembling} \label{sec-method-ens}

Ensembling \cite{breiman1996bagging} is a way to increase both quality of the model and its stability at the expense of time and computational power. In our case, we have a relative abundance of the negative class samples, as described in Section \ref{sec-data}. Hence it is possible to use different subsamples of the negative class samples for training each model in the ensemble. The specific parameters will be described in Section \ref{sec-ens}.

In the final version of our model, we settled to use mean predictions of an ensemble of six separately trained models, as a practical balance between prediction quality and execution time. Ensemble quality gain and other possible ensembling strategies are further explored in Section \ref{sec-arc-sel}.

\section{Experiments} \label{sec-exp}
\subsection{Data} \label{sec-data}

The data used for experiments was provided by a large European bank. For our experiments, we took transactional data for the clients who applied for the \textit{retail credits}. Since strict adherence to the requirements of personal data protection laws was one of the key priorities for the Bank when carrying out the project, we cannot describe our data and share it with the readers.
Furthermore, we only considered applicants who already used a debit or a credit card product in the bank. If a client has several cards, then transactions from every card was taken into account.

The available transactional data falls into subcategories: trans\-action-level (such as timestamp, country, amount, merchant type) and card-level (such as issuing branch, card type). Card-level data is duplicated verbatim for each transaction related to the corresponding card. An example of three typical card transactions is presented in Table \ref{tab-tr-data}.
We also used two derived features calculated from transactional data:
\begin{itemize}
\item difference in days between the time of current transaction and the time of previous transaction by this customer
\item time in days elapsed from the card issue date until the transaction date
\end{itemize}
Only the transactions performed before the application date are taken for training and validation.

Our training dataset represented more than 740 thousand clients with approximately 200 million transactions in total. As a target variable, we used the event of default for consumer loan during a year after its disbursement. The period of one year was selected using the performance window attribute, as described in \cite{siddiqi2005credit}.

Due to the risk of data non-stationarity, we have opted to use out-of-time validation strategy, as in \cite{KVAMME2018207}, instead of out-of-sample validation as used in \cite{khandani2010consumer} and \cite{bellotti2013forecasting}. Note that our results for the out-of-fold validation were consistently higher than that for the out-of-time validation for a range of architectures and hyperparameters, which is the common situation, as discussed in \cite{glennon2008development}.

We have used a subset of credit applications from 16-month period for training and four-month period for the out-of-time validation. Training and validation sets were the same for each considered model and baseline.

Due to a large disparity between number of positive and negative cases (because of the low default rate at the bank), we settled on the following undersampling strategy: before each experiment we selected all the positive cases and 10 times as much randomly selected negative cases. On each training epoch we used all positive cases and an equal number of negative cases, selected from the pool of negative cases.

All models in this paper where trained on the last 800 transactions for each customer when available, padding by zero was applied when the actual transaction count for a client was lower.

\subsection{Baselines}

To compare our model with other approaches, we have implemented a logistic regression based model. We have also implemented an additional model that is based on the Gradient Boosting Machine (GBM) method \cite{friedman2001greedy}.

Both logistic regression and GBM methods require a \textit{large} number of \textit{hand-crafted} aggregate features produced from the transactional data as an input to the classification model. An example of an aggregate feature would be an average spending amount in some category of merchants, such as hotels of the entire transaction history.

We used LightGBM\cite{Ke2017LightGBMAH} implementation of GBM algorithm and created nearly 7000 hand-crafted features for the application.

Similarly, for the logistic regression we manually designed about 400 aggregate features. Weight of evidence coding and binning of predictors \cite{lund2016woe} was used to transform categorical features.

\subsection{Offline execution of our method} \label{sec-exec}

\subsubsection{Encoder architecture selection} \label{sec-arc-sel}

We have experimented with a different architectures of encoders, using Long Short Term Memory (LSTM), Bidirectional Recurrent Cells \cite{schuster1997bidirectional} and Gated Recurrent Units (GRU). The results of this comparison are presented in Table \ref{tab-enc-arch}. Based of this comparison we decided to use one-layer GRU because the difference with the best performing bidirectional model was not statistically significant, while increasing complexity of the model and incurring a noticeable computational price.

\begin{table}[ht]
\caption{Encoder architecture comparison}
\begin{tabular}{ | l | c |  }
\hline
\textbf{Encoder} & \textbf{Valid ROC-AUC (STD)} \\
\hline
\textbf{GRU 1-layer} & 0.8155 (0.0015)  \\
\textbf{GRU 1-layer Bidirectional} & 0.8160 (0.0004)  \\
\textbf{LSTM 1-layer} & 0.8055 (0.0022) \\
\textbf{LSTM 1-layer Bidirectional} & 0.8058 (0.0027)  \\

\hline
\end{tabular}
\label{tab-enc-arch}
\end{table}

\subsubsection{Loss function and learning rate}

We used a batch size of 32 for the training and the batch size of 768 for validation for all the experiments. When using ranking loss, we introduced the new hyperparameter \textit{loss margin size}. We found that loss margin size of 0.1 gives the best results among all the loss hyperparameters that we tried, as shown in Table \ref{tab-loss}.

\begin{table}[ht]
\caption{Loss comparison}
\begin{tabular}{ | l | c |  }
\hline
\textbf{Loss} & \textbf{Valid ROC-AUC (STD)} \\
\hline
\textbf{BCE Loss} & 0.8124 (0.0016)  \\
\textbf{Hinge 0.5} & 0.8104 (0.0026)  \\
\textbf{Hinge 0.1} & 0.8168 (0.0017)  \\
\textbf{Hinge 0.01} & 0.8155 (0.0016)  \\
\textbf{Hinge 0.01 + BCE} & 0.8144 (0.0030)  \\
\hline
\end{tabular}
\label{tab-loss}
\end{table}

Learning rate and learning rate reduction schedule is one of the most sensitive hyperparameters which can dramatically change the performance of the model.
Note, that the optimal learning rate schedule depends heavily on loss function used, batch size and overall number of parameters in the model.
We tried several learning rates and several learning rate reduction regimes and found that for both BCE loss and ranking loss the most effective strategy was an aggressive linear learning rate reduction with gamma=0.5, as shown in Table \ref{tab-lr}. We also tried unsuccessfully instead of monotonically decreasing the learning rate to vary it cyclically as proposed in \cite{smith2017cyclical}.

\begin{table}[ht]
\caption{Learning rate schedules}
\begin{tabular}{ | l | c |  }
\hline
\textbf{Loss} & \textbf{Valid ROC-AUC (STD)} \\
\hline
\textbf{gamma = 1} & 0.8042 (0.0026)  \\
\textbf{gamma = 0.8} & 0.8144 (0.0015)  \\
\textbf{gamma = 0.5} & 0.8155 (0.0016)  \\
\textbf{gamma = 0.5, 2 cycles} & 0.8145 (0.0006)  \\
\textbf{gamma = 0.5, 3 cycles} & 0.8111 (0.0027)  \\
\hline
\end{tabular}
\label{tab-lr}
\end{table}

\subsubsection{Regularization methods}

\begin{figure}[ht]
  \caption{Regularization methods}
  \includegraphics[width=0.46\textwidth]{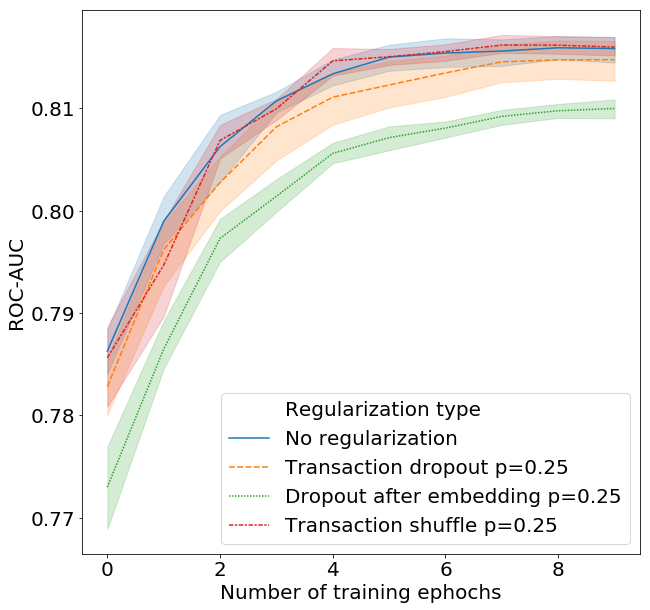}
  \label{fig-reg}
\end{figure}

Due to the low number of positive classes, all models exhibit propensity for overfitting. Therefore we tried various types of dropout regularisation, such as:
\begin{itemize}
\item \textit{Transaction dropout} that randomly drops some of the client transactions with defined probability
\item \textit{Transaction shuffle} that randomly permutes the order of client transactions
\item \textit{Dropout after embedding} that randomly zeroes some components after embedding layer
\end{itemize}
Note that, none of the aforementioned regularization methods proved effective against overfitting, as shown in Figure \ref{fig-reg}.

\subsubsection{Ensembling methods} \label{sec-ens}

\begin{figure}[ht]
  \caption{Ensemble quality comparison}
  \includegraphics[width=0.46\textwidth]{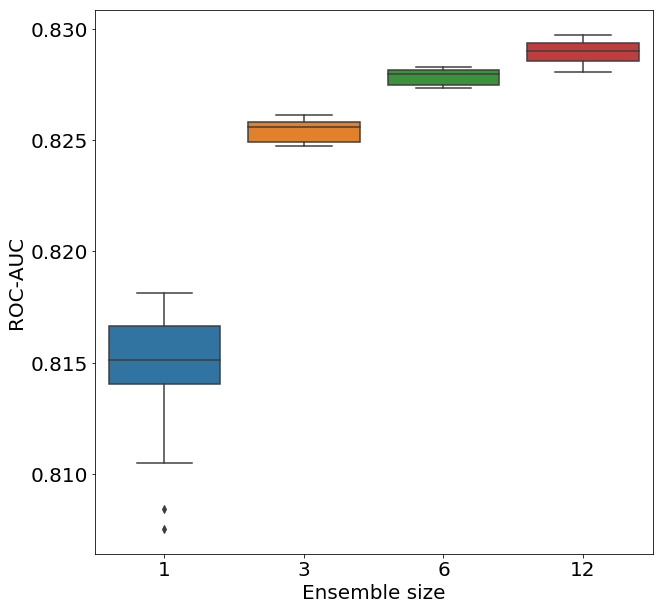}
  \label{fig-ens}
\end{figure}

We tried several different types of ensembling methods:
\begin{itemize}
\item Simple averaging of model results. Averaging predictions of different models trained with distinct negative class examples leads to both increased accuracy and reduced variabilty of results, as shown in Figure \ref{fig-ens}
\item Stochastic Weight Averaging (SWA) \cite{DBLP:journals/corr/LoshchilovH16a}.  Averaging the weights of ensemble models can significantly reduce inference time since only one model with averaged weights is used instead of the whole ensemble. But in our case averaging of weights of different models led to noticeable reduction in quality.
\item Snapshot ensembling \cite{DBLP:journals/corr/HuangLPLHW17}. Using snapshots of the same model in the final ensemble can significantly reduce training time since only one model should be trained. Unfortunately this approach does not benefit from using distinct negative class examples
\item SWA + snapshot ensembling. We found that combining SWA with snapshot ensembling for single model training by taking snapshots after a set epoch and averaging the weights leads to some reduction of variability, but the results were inconclusive and we opted for not using these advanced ensembling methods in our production model.
\end{itemize}

We opted to use a size six averaging ensemble for our production model, providing a reasonable compromise between model quality and training/inference times. As mentioned in Section \ref{sec-method-ens}, each model of the ensemble is trained on different subsamples of the negative class samples. As described in Section \ref{sec-data}, we use undersampling procedure to reduce the number of negative samples. The negative samples are selected independently for each model of the ensemble, hence each model of the ensemble is trained on slightly different subset of negative samples.

\subsection{Moving to production}

We performed massive field test of our neural scoring model in a bank's production pipeline. We used the model trained on the same dataset as discussed in Section \ref{sec-data} to pre-calculate scores for each client with a debit or credit card. Training of a full six model ensemble took about 4 hours on a Tesla P100 GPU. It took about 17 minutes to score 1 million customers on an Tesla P100 GPU. And the inference time scales linearly with the number of clients.

These scores were used to make decisions about credit applications for tens of thousands of applicants during one month and the early results are very promising. 

The potential financial gain was measured for the case if our model is used instead of the current scoring model for the applicants with enough transnational data. The preliminary financial results are measured in the millions of dollars per year, which constitutes a very significant result for the bank of this type and size.

\section{Results}

Table \ref{tab-res} presents the main results of the experiments described in Section \ref{sec-exp}.

\begin{table}[ht]
\caption{Experiment comparison}
\begin{tabular}{ | c | c | c | }
\hline
& \textbf{ROC AUC} & \textbf{N Features} \\
\hline
\textbf{Logistic regression} & 0.78 & $\sim400$ \\
\textbf{LGBM} & 0.81 & $\sim7000$ \\
\textbf{E.T.-RNN} & 0.83 & 12 \\
\hline
\end{tabular}
\label{tab-res}
\end{table}

As shown in Table \ref{tab-res}, E.T.-RNN significantly outperformed the baselines on our data. Moreover, one of the crucial features of our approach is that we did not have to do feature engineering for our method, unlike the classical methods which rely heavily on the hand-crafted features (e. g. 400 features for Logistic regression and 7000 features for LGBM).

\subsection{Training dataset size}

\begin{figure}[ht]
  \caption{E.T.-RNN has steeper learning curve than LGBM.}
  \includegraphics[width=0.46\textwidth]{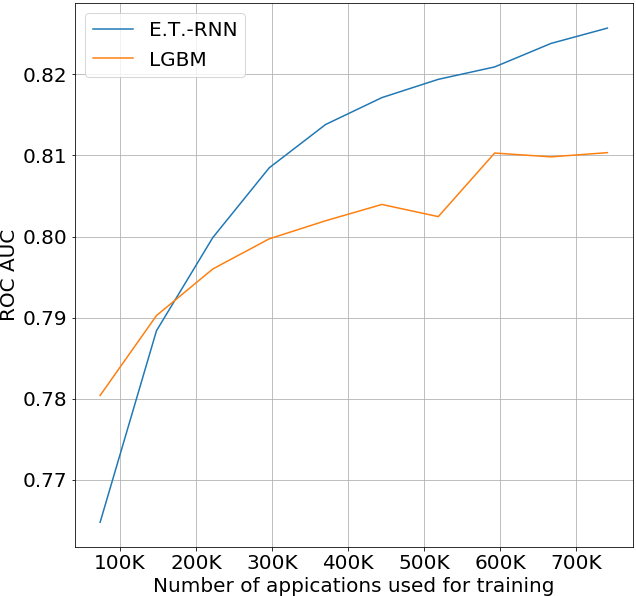}
  \label{fig-lc}
\end{figure}

Note that the results presented in Table \ref{tab-res} were achieved on the full dataset described in Section \ref{sec-data}. We also conducted a series of experiments to estimate model performance for different dataset sizes.
As Figure \ref{fig-lc}, shows LGBM outperforms our approach for \textit{small} volumes of data, as measured in terms of the number of applications (on the X axis). However, given enough data, E.T.-RNN method significantly outperforms the classical approaches. This observation is in line with the well-known understanding that neural networks outperform classical methods on large datasets.

Also note that E.T.-RNN has steeper learning curve than LGBM. Hence the performance gap would increase even further with more available data.

\subsection{Transaction count}

\begin{figure}[ht]
  \caption{Classification quality vs number of transactions}
  \includegraphics[width=0.46\textwidth]{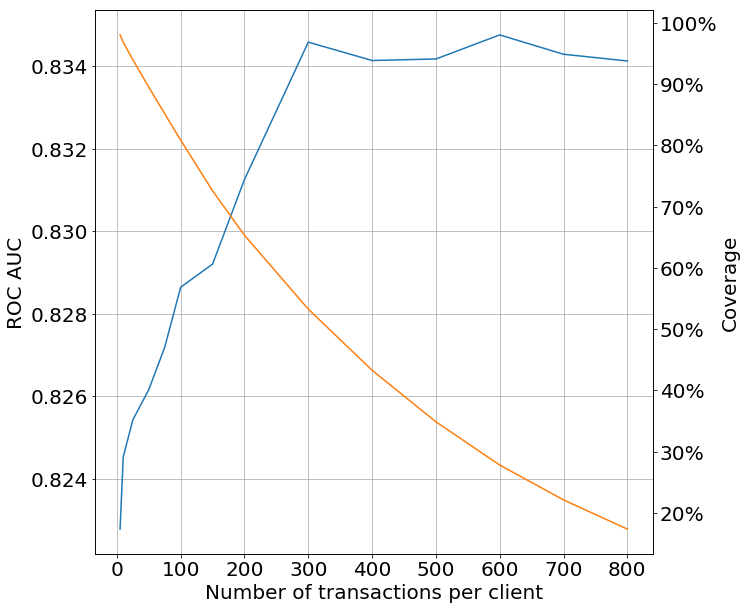}
  \label{fig-tc}
\end{figure}

Performance of our model depends heavily on the number of available transactions per client. As Figure \ref{fig-tc} shows, scoring quality increases untill we reach around 350 transactions. Beyond this level, performance increase due to additional transactions is insignificant enough to be overshadowed by statistical variations in the data. Furthermore the share of clients having more than 350 transactions is about 50 percent for our dataset. This means that our model achieves significant hit rate when scoring clients of the bank. On the other hand, our method is still effective even for the applicants with a low number of transactions. For clients with more than 25 transactions (about 95 percent of total number of clients), we reach 82.5 ROC-AUC.

\begin{figure}[ht]
  \caption{Classification quality for customers grouped by number of transactions}
  \includegraphics[width=0.46\textwidth]{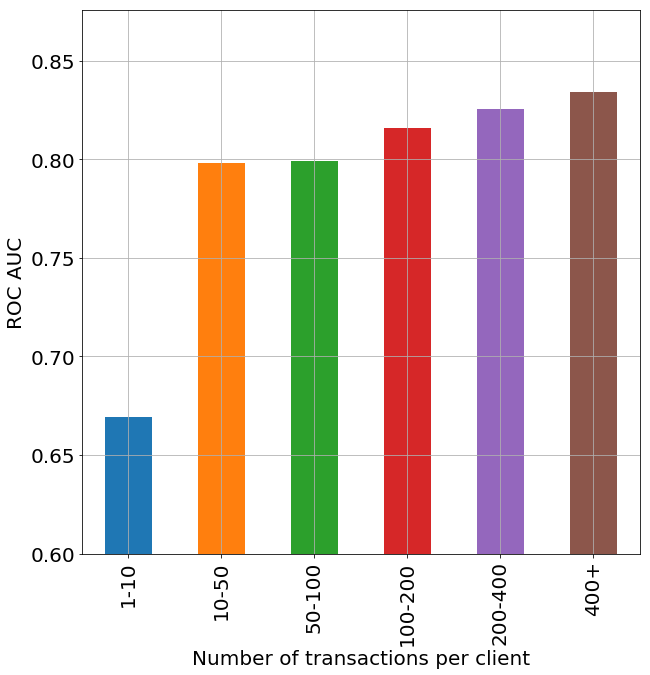}
  \label{fig-tcb}
\end{figure}
\section{Discussion}

Our method worked well for the following reasons:
\begin{itemize}
    \item Reasonably large number of customers in training dataset. Neural networks have lots of learnable parameters comparing to the classical approaches and, hence, require more data than classical methods. This is also true in our case as presented on Figure \ref{fig-lc}.
    \item Low-level, granular data. Our data can be described as a series of events and each event consists of several variables. Note that if data structure is relatively simple, our method may not work better than the traditional approaches. For example, for the data from the application questionnaire there is no need for sophisticated neural network models. Even classical ML approaches, like logistic regression would work reasonably well on the data with simple table-like structure.
    \item High-frequency data (as discussed before, more than 80 percents of customers have at least 100 transactions.
\end{itemize}

Our method worked because we applied sophisticated neural network method (as discussed in Section \ref{sec-exec} we tried numerous other DL-based approaches, and many of them did not work that well) on the data exhibiting aforementioned characteristics.

To summarize, our E.T.-RNN approach would possibly work better than classical methods in cases where data is in low-level, granular form and there is enough data to train complex neural net based model.

\section{Conclusions}

In this paper we proposed a novel E.T.-RNN method which allows to use fine-grained transactional data for credit scoring.
We tested our method against the benchmarks on the historical data and achieved superior performance. We also conducted a pilot study on banking customers and produced significant financial gains for the bank.

The significant advantage of our approach is that even complex multivariate time-series data can be directly used for training without any need for feature design. As was demonstrated in \cite{erhan2009visualizing}, the neural network learns meaningful internal representations of the input data during training, and this drastically reduces the need to generate hundreds or even thousands of hand-crafted aggregate features, as is typically done in credit scoring applications. This means that our method does not require any significant domain-specific expertise for feature design.
Also, our model works exclusively on the transactional data and therefore does not require any additional input from the client that means that we can make credit loan decisions very fast, ideally in nearly real-time, because the whole credit scoring process is fully automated.
Moreover information in the transactional data is exceptionally hard to forge. Hence, there is no need of costly checks for the correctness of such data, unlike data provided by the client or obtained from some other sources.
Still another advantage of our method is that even a person without any credit history can be reliably accessed for credit-worthiness, his or her transactional history constituting a source for estimating credit risks.
Finally, this method provides a \textit{fair} approach to credit decision making because it \textit{does not} rely on personal demographic information of an individual and, therefore, cannot discriminate applicants based on various demographic factors.
For all these reasons, we believe that the proposed credit scoring approach has a potential to disrupt current loan practices in the retail banking industry.

One issue with our method is lack of interpretability. Neural networks constitute black-box models by their nature. The ability to produce rich models on top of raw data representation is the main strength of neural networks. But this ability also leads to significant interpretability problem, which is the main weakness of complex models. Also note that this issue is applicable not only to our method, but also to most other advanced machine learning methods, such as  Gradient Boosting Machine, since they also suffer from the lack of interpretability.

Different organizations around the worlds have different philosophies regarding applying black-box models in credit scoring. In some countries, lack of interpertablilty is considered less appropriate while in other parts of the world it is considered more appropriate to do so. We believe that this issue will be less relevant moving forward because of the significant progress in solving the black-box interpretetaion problem, including that of neural networks, that have been achieved over the past few years. \cite{DBLP:journals/corr/ChoiBSSS16}, \cite{gupta2018lisa}, \cite{mccoy2018rnns} constitute some examples of the recent work. Based of this progress, the black-box interpretation problem should be successfully addressed moving forward.

As a future work, we plan to study more effective method of regularization, which would allow us to use the data we have available more effectively.
Furthermore, we plan to focus on even more effective ways to integrate time into our model. In particular our model is not sensitive to shifting all the customers transactions in time (e.g. shifting by one month back), and we plan to work on this problem.
Finally we also plan to work on other types of loans, such as mortgage loans, which differ from retail loans in several respects.

\bibliographystyle{ACM-Reference-Format}
\bibliography{sigconf}

\end{document}